\documentclass[conference]{IEEEtran}
\IEEEoverridecommandlockouts
\usepackage{textcomp}

\usepackage{blkarray}

\usepackage{fancyhdr}
\usepackage{blkarray}
\usepackage{amsmath,amssymb,amsfonts}
\usepackage{graphicx}
\usepackage{xcolor}
\usepackage{booktabs}
\usepackage{array}
\usepackage{multirow}
\usepackage{float}

\usepackage[numbers,sort&compress]{natbib}  

\usepackage{algorithm}              
\usepackage{algpseudocode}          

\usepackage[hidelinks]{hyperref}
\usepackage{xurl}

\def\BibTeX{{\rm B\kern-.05em{\sc i\kern-.025em b}\kern-.08em
    T\kern-.1667em\lower.7ex\hbox{E}\kern-.125emX}}
\begin{document}

\title{B\text{-}TGAT: A Bi-directional Temporal Graph Attention Transformer for Clustering Multivariate Spatiotemporal Data\\
\thanks{This work is supported by HDR Institute: HARP - Harnessing
Data and Model Revolution in the Polar Regions (OAC-2118285).}
}

\author{\IEEEauthorblockN{1\textsuperscript{st} Francis Ndikum Nji}
\IEEEauthorblockA{\textit{Dept of Information Systems} \\
\textit{University of Maryland, Baltimore County}\\
Baltimore, USA \\
\email{fnji1@umbc.edu}}
\and
\IEEEauthorblockN{2\textsuperscript{nd} Vandana Janeja}
\IEEEauthorblockA{\textit{Dept of Information Systems} \\
\textit{University of Maryland, Baltimore County}\\
Baltimore, USA \\
vjaneja@umbc.edu}
\and
\IEEEauthorblockN{3\textsuperscript{rd} Jianwu Wang}
\IEEEauthorblockA{\textit{Dept of Information Systems} \\
\textit{University of Maryland, Baltimore County}\\
Baltimore, USA \\
jianwu@umbc.edu}}

\author{
\IEEEauthorblockN{Francis Ndikum Nji}
\IEEEauthorblockA{Information Systems\\ UMBC, Baltimore, USA\\ \texttt{fnji1@umbc.edu}}
\and
\IEEEauthorblockN{Vandana Janeja}
\IEEEauthorblockA{Information Systems\\ UMBC, Baltimore, USA\\ \texttt{vjaneja@umbc.edu}}
\and
\IEEEauthorblockN{Jianwu Wang}
\IEEEauthorblockA{Information Systems\\ UMBC, Baltimore, USA\\ \texttt{jianwu@umbc.edu}}
}

\maketitle

\begin{abstract}
Clustering high-dimensional multivariate spatiotemporal climate data is challenging due to complex temporal dependencies, evolving spatial interactions, and non-stationary dynamics. Conventional clustering methods, including recurrent and convolutional models, often struggle to capture both local and global temporal relationships while preserving spatial context. We present a time distributed hybrid U-Net autoencoder that integrates a Bi-directional Temporal Graph Attention Transformer (B\text{-}TGAT) to guide efficient temporal clustering of multidimensional spatiotemporal climate datasets. The encoder and decoder are equipped with ConvLSTM2D modules that extract joint spatial–temporal features by modeling localized dynamics and spatial correlations over time, and skip connections that preserve multiscale spatial details during feature compression and reconstruction. At the bottleneck, B\text{-}TGAT integrates graph-based spatial modeling with attention-driven temporal encoding, enabling adaptive weighting of temporal neighbors and capturing both short- and long-range dependencies across regions. This architecture produces discriminative latent embeddings optimized for clustering. Experiments on 3 distinct spatiotemporal climate datasets demonstrate superior cluster separability, temporal stability, and alignment with known climate transitions compared to state-of-the-art baselines. The integration of ConvLSTM2D, U-Net skip connections, and B\text{-}TGAT enhances temporal clustering performance while providing interpretable insights into complex spatiotemporal variability, advancing both methodological development and climate science applications.
\end{abstract}


\section{Introduction}
\label{sec:intro}
Spatiotemporal data plays a central role in numerous scientific disciplines, particularly in Earth sciences \cite{gorelick2017google}, atmospheric research \cite{ERA5}, and climate science \cite{lee2023synthesis}. Such datasets are inherently high-dimensional and multivariate, often spanning four dimensions that include time, latitude, longitude, and a wide range of measured climate variables such as temperature, humidity, and precipitation. With the increasing availability of high-resolution reanalysis products and satellite observations, researchers are now confronted with the challenge of extracting meaningful structure from data that is both voluminous and complex. One of the most effective ways to analyze these datasets is through clustering, where similar patterns are grouped together to reveal hidden structures, temporal regimes, and transitions across spatial and temporal scales. However, clustering high-dimensional spatiotemporal climate data remains a difficult problem due to the combination of nonlinear dependencies, dynamic interactions across spatial regions, and evolving temporal patterns.

Traditional clustering algorithms such as k-means and Gaussian mixture models were primarily designed for static, low-dimensional tabular data and therefore struggle when applied directly to spatiotemporal data. A naive reduction of 4D climate data into two-dimensional tabular representations
inevitably leads to both a loss of structural information \citep{LeCun1998, Shi2015}
and prohibitively high dimensionality \citep{Strommen2023,Christiansen2021}. While dimensionality reduction techniques such as principal component analysis (PCA)\cite{JolliffeCadima2016}  have been used to mitigate this challenge, these methods impose linear assumptions that fail to capture the nonlinear dependencies intrinsic to climate systems. Furthermore, the reduction process often overlooks critical spatiotemporal patterns that are essential for identifying meaningful clusters. As a result, clustering performance using such methods is often suboptimal and limited in its ability to reveal the underlying variability in climate dynamics.

Recent advances in machine learning and deep learning have introduced new opportunities to address the shortcomings of conventional methods. Deep autoencoders, convolutional neural networks (CNNs), and recurrent neural networks (RNNs) have been successfully applied to climate and geoscience problems, enabling the extraction of latent representations that preserve nonlinear relationships in the data. CNN-based approaches excel at capturing spatial correlations \cite{kim2016accurate, goodfellow2016deep}, while recurrent models such as LSTMs are capable of modeling temporal dependencies \cite{graves2012supervised}. Nevertheless, methods that rely solely on either convolutional or recurrent architectures do not fully capture the joint spatiotemporal dynamics present in climate systems. Moreover, many existing models lack mechanisms for explicitly handling evolving spatial interactions over time, which are critical in multivariate spatiotemporal climate datasets where dependencies between variables and regions change dynamically.

To address these challenges, we present a novel hybrid deep learning framework for temporal clustering of high-dimensional spatiotemporal climate data. Our model is built upon a time-distributed U-Net autoencoder architecture that leverages ConvLSTM2D modules to extract joint spatial-temporal features. ConvLSTM2D layers are well-suited for this task as they incorporate convolutional operations within the recurrent LSTM structure, enabling simultaneous modeling of localized spatial dependencies and temporal dynamics. The encoder progressively compresses the input data into lower-dimensional latent representations while preserving multiscale spatial details through skip connections that link encoder and decoder layers. The decoder reconstructs the input from these latent features, providing a dual objective of minimizing reconstruction error while also optimizing latent embeddings for clustering. This autoencoding structure ensures that the model not only captures meaningful latent representations but also maintains consistency with the original high-dimensional spatiotemporal input.

At the bottleneck of the U-Net autoencoder, we integrate a Bi-directional Temporal Graph Attention Transformer (B\text{-}TGAT) to further enhance the representation of temporal dependencies across spatial regions. Unlike traditional RNN-based methods, B\text{-}TGAT leverages graph-based spatial modeling and attention-driven temporal encoding to assign adaptive weights to temporal neighbors. This mechanism allows the model to effectively capture both short- and long-range dependencies across space and time, an essential capability for climate data where local phenomena can interact with global teleconnections. By embedding spatiotemporal data within a graph structure, B\text{-}TGAT provides an interpretable and flexible mechanism for modeling evolving interactions across climate variables and regions. This integration ensures that the resulting latent embeddings are both discriminative and informative, thereby enabling robust clustering.

The final stage of the model incorporates a custom clustering layer applied to the latent embeddings generated by the B\text{-}TGAT bottleneck to generate clustering results. The clustering layer uses the inherent logic of the Student’s t-distribution and iteratively improves the result. At the same time, the decoder module adjusts its weights to reduce the disparity between the input and reconstructed data while learning to reconstruct the high-dimensional input data from lower-dimensional latent features. Simultaneously, the autoencoder structure enforces a reconstruction loss that minimizes the disparity between the original input and its reconstruction. The combination of clustering and reconstruction losses enables the model to perform joint optimization, producing latent representations that are both low-dimensional and optimized for clustering accuracy. This dual-objective approach ensures that the learned latent space is not only compact and representative but also directly aligned with the clustering task.

We validate our framework on three distinct multivariate spatiotemporal climate datasets, which represent different spatiotemporal scales and climate variables. Experimental results demonstrate that our model consistently outperforms state-of-the-art baselines in terms of cluster separability, temporal stability, and alignment with known climate transitions. By effectively integrating ConvLSTM2D layers, U-Net style autoencoder with skip connections, and B\text{-}TGAT into a unified framework, our approach achieves superior performance in our temporal clustering tasks while providing interpretable insights into the complex dynamics of climate variability. These insights are particularly valuable for advancing climate science applications, such as identifying regions undergoing rapid transitions, detecting emergent teleconnections, and improving understanding of nonlinear climate processes.

In summary, this work contributes a novel end-to-end deep learning framework for temporal clustering that addresses the unique challenges of multidimensional spatiotemporal data clustering with data transformation from 4D multidimensionality to 2D high dimensionality. The primary contributions of this study are threefold: first, the design of a hybrid U-Net autoencoder architecture with stacked ConvLSTM2D modules for joint spatial–temporal representation learning; second, the integration of a B\text{-}TGAT layer at the bottleneck to capture evolving spatial interactions and attention-based temporal dependencies at both local and global scales; and third, the development of a temporal clustering pipeline that jointly optimizes latent representations using both clustering and reconstruction losses. By bridging advances in deep representation learning with clustering, this framework advances methodological development in unsupervised spatiotemporal learning and contributes to improved interpretability in climate science applications.
Our implementation code is publicly available
The remainder of the paper is structured as follows. Section \ref{bknrw} summarizes the background and state-of-the-art related works while Section \ref{sec:prob_st} describes the problem in detail. Section \ref{sec:method} presents our proposed solution while Section \ref{sec:Evaluation} discusses how our model is evaluated with experiment results and Section \ref{concl} concludes our research.

\section{Background and Related Work} \label{bknrw}
\subsection{Background}\label{sec:Background}
Earth’s climate system is a highly complex, nonlinear, and interconnected global system composed of multiple dynamic components such as atmospheric temperature, ocean heat content, Arctic sea ice, precipitation, wind circulation, surface pressure, aerosol concentrations, and cloud dynamics. Climate change has emerged as one of the most pressing global challenges \cite{lee2024climate} due to its profound influence on ecosystems, human societies, and the frequency and intensity of extreme weather events \cite{portner2022summary}. The inherent complexity of the climate system arises not only from the diversity of its components but also from their nonlinear interactions and feedback mechanisms across spatial and temporal scales. Understanding these interactions and the emergent variability they produce remains a fundamental task for climate science. Among the key components of Earth’s climate are the coupled interactions between the atmosphere, ocean, cryosphere, and biosphere. Processes such as air-sea exchange \cite{taylor2018increasing}, ice-albedo feedback \cite{riihela2021recent}, and cloud-radiation interaction \cite{bony2015clouds} create tightly coupled feedback loops that regulate energy balance and circulation patterns. For example, marine boundary layer clouds exert strong radiative forcing by reflecting incoming solar radiation, and their microphysical properties are sensitive to both sea surface temperature and atmospheric aerosol concentration. Similarly, Arctic sea ice plays a pivotal role in modulating polar amplification by influencing ocean-atmosphere heat exchange and altering planetary albedo. These feedbacks operate across a wide range of spatiotemporal scales, from local synoptic disturbances to decadal climate variability, leading to complex system-wide behavior that is difficult to model and interpret directly.

Observational evidence indicates that the variability of these interactions manifests across broad temporal horizons, ranging from daily synoptic events to multi-decadal climate oscillations. This variability is spatially heterogeneous, driven by local forcing mechanisms but often coupled with global teleconnections such as the El Niño-Southern Oscillation (ENSO) \cite{trenberth2019nino}, the North Atlantic Oscillation (NAO) \cite{pinto2012past}, and Arctic Oscillation (AO) \cite{he2017impact}. Quantifying how different components contribute to such variability requires analyzing datasets that span wide latitudinal and longitudinal ranges and extend over long temporal periods. Such datasets are inherently high-dimensional, containing measurements across space, time, and multiple variables, which makes them computationally expensive to process and challenging to interpret without advanced analytical methods.

The scale and complexity of spatiotemporal climate datasets necessitate approaches that can reduce problem size while preserving critical structural patterns. One effective strategy is to partition the data into groups of spatially and temporally coherent regimes. By clustering observations based on similarity in their spatial and temporal attributes, researchers can simplify the analysis of large-scale climate processes. Each cluster represents a coherent regime of variability, within which small-scale perturbations are more interpretable and boundaries of global-scale effects become clearer. This allows researchers to uncover latent relationships between variables and identify transition patterns across different climate states.

Formally, the problem of studying atmospheric and climate interactions can be formulated as an unsupervised clustering task on four-dimensional datasets characterized by time, longitude, latitude, and measured variables. Unlike supervised learning problems, climate clustering lacks explicit labels, requiring the development of methods that can extract meaningful structures directly from the data. Moreover, the clustering algorithm must account for both spatial similarity and temporal evolution, ensuring that the generated groups not only capture co-located features but also preserve the dynamic dependencies over time. This makes spatiotemporal clustering a particularly suitable framework for disentangling the complexity of climate dynamics and providing interpretable insights into the variability of the Earth system.

The challenge of clustering spatiotemporal data arises from the interplay between high dimensionality and the preservation of nonlinear dependencies. Climate datasets, such as those derived from reanalysis products like ERA5 \cite{ERA5} or satellite observations, typically contain millions of grid points across both space and time, resulting in an immense feature space. Each grid point evolves over time, interacting with neighboring regions and responding to large-scale climate drivers such as El Niño–Southern Oscillation or Arctic amplification. Capturing these patterns require methods that can simultaneously account for spatial heterogeneity and temporal variability.



\subsection{Related Work}\label{sec:related}
\noindent \textbf{{Traditional Clustering Algorithms}}: Unsupervised clustering is one of the most extensively studied problems in machine learning, with classical approaches focusing on feature selection, similarity measures, grouping strategies, and cluster validation. Among the most popular methods, the k-means algorithm \cite{lloyd1982least} remains widely applied due to its computational efficiency and scalability. However, k-means is well-suited primarily for low-dimensional data and struggles with high-dimensional spatiotemporal datasets, where nonlinear dependencies and temporal correlations are critical. Another widely used approach is Density-Based Spatial Clustering of Applications with Noise (DBSCAN) \cite{ester1996density}, which automatically identifies clusters of arbitrary shape based on neighborhood density. Despite its strengths, DBSCAN performs poorly when data density varies significantly across regions, which is often the case in multivariate climate datasets. Hierarchical clustering \cite{johnson1967hierarchical}, particularly the agglomerative variant, provides another perspective by recursively merging data points into larger groups. While this approach is conceptually simple, it suffers from high computational complexity and scalability issues when applied to large, high-dimensional datasets.  

To extend the applicability of these traditional methods to spatiotemporal problems, several works have incorporated dimensionality reduction techniques such as principal component analysis (PCA) and independent component analysis (ICA) \cite{jolliffe2002principal}. While these linear transformations reduce the dimensionality of the data before clustering, they fail to capture nonlinear and dynamic spatiotemporal relationships inherent in climate data. As a result, traditional clustering methods, while foundational, fall short of effectively analyzing large-scale, high-dimensional datasets that require integration of spatial and temporal dependencies.

\noindent \textbf{{Deep Learning–based Clustering}}: The limitations of traditional clustering methods have motivated the development of deep learning–based approaches, which are better equipped to model nonlinear, high-dimensional data. Deep Embedded Clustering (DEC) \cite{xie2016unsupervised} introduced the paradigm of jointly learning representations and cluster assignments by minimizing a Kullback–Leibler (KL) divergence loss between predicted and target distributions. Extensions of DEC and related autoencoder-based methods have been applied to time series data, though many approaches either focus solely on temporal patterns or image-level spatial similarities, neglecting the joint spatiotemporal structure.

To address this gap, spatiotemporal autoencoders have emerged, combining convolutional neural networks (CNNs) with recurrent architectures such as Long Short-Term Memory (LSTM) networks. In particular, ConvLSTM2D \cite{shi2015convlstm} has proven highly effective for spatiotemporal data, as it integrates convolutional operations into recurrent units, allowing the model to simultaneously capture localized spatial patterns and their evolution over time. ConvLSTM2D has been successfully applied to precipitation nowcasting \cite{shi2017deep}, sea ice prediction in the Arctic \cite{wang2019deep}, and regional climate variability detection \cite{liu2020spatiotemporal}, demonstrating its capability to extract meaningful representations from complex climate datasets. These applications highlight the model’s ability to capture both fine-scale spatial correlations and their evolution across temporal sequences. Nevertheless, ConvLSTM2D alone is limited in its ability to adaptively capture long-range temporal dependencies and evolving spatial interactions, particularly in datasets where teleconnections and non-stationary dynamics dominate.

Recent advances in graph neural networks (GNNs) have opened new opportunities for modeling dynamic spatial and temporal relationships. Temporal Graph Attention Networks (TGAT) \cite{xu2020inductive} extend graph convolution by incorporating temporal attention, enabling models to learn evolving spatiotemporal interactions while adaptively weighting neighbors based on their temporal relevance. In geoscience, GNNs and TGAT variants have been explored for applications such as air quality forecasting \cite{jiang2023graph}, urban climate modeling \cite{li2023spatiotemporal}, and climate teleconnection discovery \cite{peng2021graph}. These studies underscore the potential of TGAT in domains where spatial structures evolve with time and causal dependencies span multiple scales. While applications of TGAT in climate science are still limited, their ability to integrate graph-based spatial modeling with attention-driven temporal encoding makes them well-suited for spatiotemporal clustering tasks.

\section{Problem Definition}\label{sec:prob_st}
The goal of the proposed model is to accurately assign data points unto different clusters based on the latent spatial and temporal features learned by the deep autoencoder model. Let us assume that $n$ atmospheric variables $(x_i)$ are measured over a grid region covering $L$ longitudes and $W$ latitudes and stored in a vector \(X = \{x_1, x_2, x_3, \ldots, x_n\}.\)
Thus, for each time step, every grid location has $n$ values corresponding to all variables. These variables are measured for $T$ different time steps, such that \(X_i = \{x_1, x_2, x_3, \ldots, x_n\}, \quad i \in \{1, \ldots, T\}.\)

\textbf{Input:} \( Dataset = \{X_1, X_2, X_3, ..., X_T\}\),
\begin{equation}
\begin{split}
X_{i} & =\begin{Bmatrix}
\begin{bmatrix}
 x_{1}(1,1)&x_{1}(1,2)  & \cdots  & x_{1}(1,W) \\ 
 x_{1}(2,1)&x_{1}(2,2)  & \cdots  & x_{1}(2,W) \\ 
 \vdots &  \vdots  & \ddots  &\vdots  \\ 
 x_{1}(L,1)&x_{1}(L,2)  & \cdots  & x_{1}(L,W)
\end{bmatrix} \vspace{0.2cm}\\ 
\begin{bmatrix}
 x_{2}(1,1)&x_{2}(1,2)  & \cdots  & x_{2}(1,W) \\ 
 x_{2}(2,1)&x_{2}(2,2)  & \cdots  & x_{2}(2,W) \\ 
 \vdots &  \vdots  & \ddots  &\vdots  \\ 
 x_{2}(L,1)&x_{2}(L,2)  & \cdots  & x_{2}(L,W)
\end{bmatrix} \\
 \vdots \\
\begin{bmatrix}
 x_{n}(1,1)&x_{n}(1,2)  & \cdots  & x_{n}(1,W) \\ 
 x_{n}(2,1)&x_{n}(2,2)  & \cdots  & x_{n}(2,W) \\ 
 \vdots &  \vdots  & \ddots  &\vdots  \\ 
 x_{n}(L,1)&x_{n}(L,2)  & \cdots  & x_{n}(L,W)
\end{bmatrix} 
\end{Bmatrix}
\end{split}
\end{equation}
\noindent Where \(X_i\) represents one observation, \(x\) represents one variable of an observation, \(i \in \{1, ..., T\}\), \(T\) represents the number of time steps, \(n\) represents the number of atmospheric variables, \(L\) and \(W\) represent the longitude and latitude respectively.

\textbf{Output:} Our proposed clustering model should partition the \(Dataset = \{X_1, X_2,..., X_T\}\) into \(k\) clusters: \(C_1, C_2, \dots, C_k\), where \(k<T\), such that objects within the same cluster are similar to each other and dissimilar to those in other clusters. Formally:\begin{equation*}
\begin{split}
    C_1=\{X_{C_1}^1, \: X_{C_1}^2,\: & ..., X_{C_1}^{n_1}\}, \: C_2=\{X_{C_2}^1, \: X_{C_2}^2, ..., X_{C_2}^{n_2}\}, \\
    &C_k=\{X_{C_k}^1, \: X_{C_k}^2, ..., X_{C_k}^{n_k}\} 
\end{split}    
\end{equation*}
\begin{equation*}
X_{C_j}^{i} \in X, \: i\in \{1, ..., n_j\}, \: j\in \{1, ..., k\} 
\end{equation*} 
\noindent , where \(n_j\) = number of observations of cluster \(j\), and the union of all clusters covers the entire dataset:
\begin{equation*}
\bigcup_{j=1}^{k}C_j=X \textrm{ and } C_j\cap C_l=\varnothing
\end{equation*}
Here, \(j\neq l\) and \((j, l)\in \{1, ..., k\}\) for each pairs of clusters.

\section{Proposed Methodology}\label{sec:method}

This section details the proposed end-to-end B\text{-}TGAT framework that integrates a time-distributed ConvLSTM2D U-Net autoencoder with a Bi-directional Temporal Graph Attention Transformer (B\text{-}TGAT) bottleneck and a clustering head optimized by a joint clustering-reconstruction objective. The framework is designed to learn compact, discriminative latent representations from complex, nonlinear, multivariate spatiotemporal climate data and to partition the temporal observations into coherent regimes. \\


\begin{figure*}[htp]
    \centering
    \includegraphics[width=0.99\textwidth]{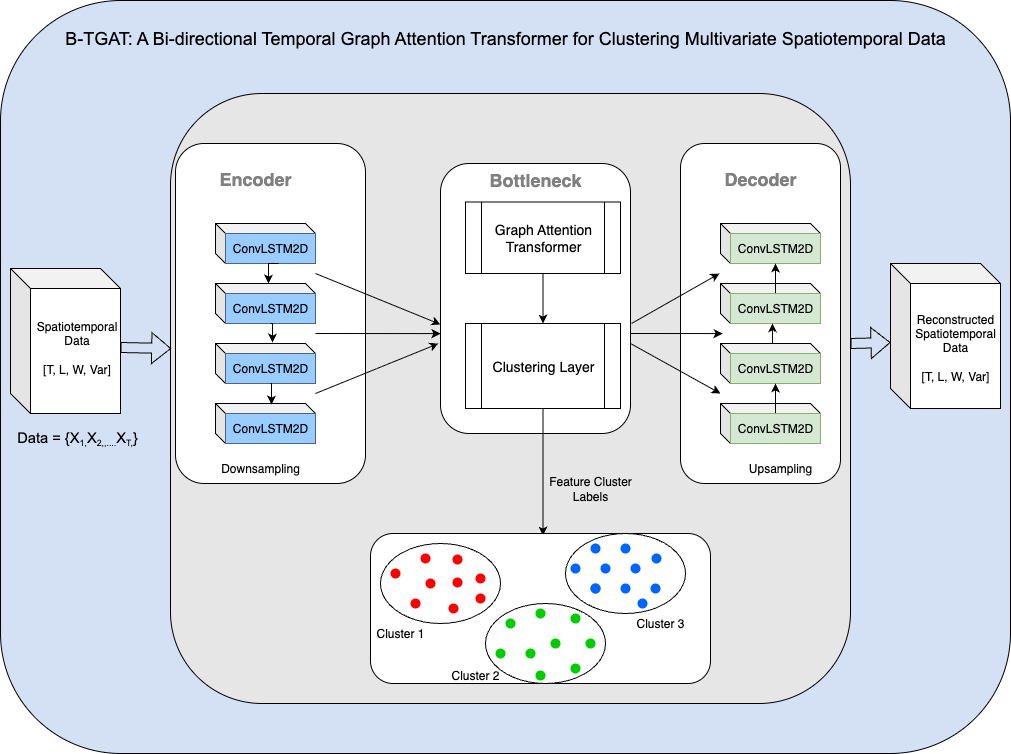}
    \caption{\textbf{Architecture of our proposed Bi-directional Temporal Graph Attention Transformer (B\text{-}TGAT)} model: 
}
    \label{fig:j}
\end{figure*}

\subsection{{Encoder: Multiscale Representation with ConvLSTM2D}:} 

Let the input to the encoder be a five-dimensional tensor \( \mathcal{X} \in \mathbb{R}^{B \times T \times H \times W \times C} \), where: \( B \): Batch size, \( T \): Temporal length (e.g., number of days for daily data), \( H, W \): Height and width (spatial resolution), \( C \): Number of climate variables per grid cell (e.g., temperature, snow depth, etc.). The encoding phase follows a U-Net downsampling structure composed of stacked \textit{TimeDistributed ConvLSTM2D} layers with spatial max pooling. The encoder extracts spatiotemporal features via stacked ConvLSTM2D blocks that operate on the full sequence. Each block applies convolutional recurrent units with $3 \times 3$ kernels to capture localized spatial patterns while propagating temporal states across frames. For an intermediate feature tensor $X^{(\ell)} \in \mathbb{R}^{T \times H_\ell \times W_\ell \times C_\ell}$ at level $\ell$, ConvLSTM2D produces \(X^{(\ell)}_{\text{conv}} = \mathrm{ConvLSTM2D}_{\ell}\!\left( X^{(\ell)} \right)\), \(X^{(\ell)}_{\text{norm}} = \mathrm{LayerNorm}\!\left( X^{(\ell)}_{\text{conv}} \right)\), followed by a residual projection to stabilize gradients and preserve information across depth. Spatial downsampling uses $\mathrm{MaxPooling3D}$, reducing $(H_\ell,W_\ell)$ while keeping the temporal resolution $T$ intact, yielding a multiscale pyramid $\{X^{(1)}, X^{(2)}, X^{(3)}, X^{(4)}\}$ with channel capacities $\{64, 128, 256, 512\}$. This hierarchy allows the encoder to summarize fine-scale texture and mesoscale structures alongside their temporal evolution, which is crucial for climate fields exhibiting scale-interacting dynamics and non-stationarity.

To mitigate information loss during progressive compression, the encoder establishes U-Net style skip connections from each resolution level to its symmetric decoder level. Let $S^{(\ell)}$ denote the skip tensor saved from encoder level $\ell$. These pathways re-inject high-frequency spatial details and boundary cues into the decoder, improving reconstruction fidelity and encouraging the latent code to capture regime-defining structure rather than expend capacity on pixel-level memorization. The skip topology also facilitates stable end-to-end optimization by shortening gradient paths and reducing vanishing signal in long temporal sequences.

At the deepest level, the model converts spatial feature maps into a graph sequence and applies a B\text{-}TGAT-style bottleneck to capture adaptive, long-range temporal dependencies coupled with spatial relations. Given $X^{(4)} \in \mathbb{R}^{T \times H_4 \times W_4 \times C_4}$, we reshape each time slice into a node feature matrix with $N=H_4W_4$ nodes and $F=C_4$ features: $Z_t \in \mathbb{R}^{N \times F}$. A graph $\mathcal{G}_t=(\mathcal{V},\mathcal{E}_t)$ is formed at each $t$ by connecting nodes based on spatial adjacency through $k$NN in feature space, producing a sparse adjacency matrix $\mathbf{A}_t$. A single-head graph attention layer computes
\[
\tilde{Z}_t = \mathrm{softmax}\!\left(\phi(Z_t)\, \mathbf{a}\, \phi(Z_t)^\top \right) Z_t,
\]
where $\phi(\cdot)$ is a learned linear map and $\mathbf{a}$ parameterizes pairwise compatibility. This operation adaptively weights contributions from spatial neighbors, allowing the representation to emphasize dynamically relevant structures. To encode temporal dynamics beyond Markovian recurrency, we stack the graph-refined embeddings across time, obtaining $G \in \mathbb{R}^{T \times F'}$ after global pooling over nodes. A bidirectional temporal encoder $\mathrm{BiLSTM}$ summarizes the sequence,
\[
b = \mathrm{BiLSTM}(G) \in \mathbb{R}^{d_b},
\]
capturing forward and backward temporal context. This B\text{-}TGAT bottleneck functions as a temporal abstraction layer that fuses local spatial reasoning (via graph attention) with sequence-level temporal integration, producing a compact code that is both spatially aware and temporally discriminative. Finally, a dense projection maps $b$ to the latent vector $E \in \mathbb{R}^{d}$ used by the decoder and clustering head.\\

\subsection{{Decoder: Time-Consistent Reconstruction}:}

The decoder mirrors the encoder by expanding the latent code $E$ into a spatiotemporal tensor and then applying a sequence of $\mathrm{UpSampling3D}$ and ConvLSTM2D blocks. At each scale, the decoder concatenates the upsampled representation with the corresponding skip tensor $S^{(\ell)}$ to recover the spatial detail suppressed during encoding. Let $\widehat{X}^{(\ell)}$ be the decoder feature at level $\ell$; the fusion is \(\widehat{X}^{(\ell)}_{\text{fuse}} = \mathrm{Concat}\!\left(\widehat{X}^{(\ell)},\, S^{(\ell)}\right),\) \quad \(\widehat{X}^{(\ell)} = \mathrm{ConvLSTM2D}_{\ell}\!\left(\widehat{X}^{(\ell)}_{\text{fuse}}\right),\) propagating temporal coherence while restoring multiscale spatial structure. The reconstruction head is a final ConvLSTM2D mapping to $n$ channels with $\tanh$ activation, delivering the sequence $\{\widehat{X}_t\}_{t=1}^{T}$. Maintaining recurrent decoding ensures that temporal phase information is preserved, which is essential for discriminating regimes that differ in evolution rather than in single-frame appearance.\\

\subsection{{Clustering Assignment: Latent Regime Discovery with Self-Refining Targets}:} 

The clustering layer is designed to assign the latent representations produced by the encoder into $k$ clusters in an unsupervised manner. At the beginning of training, the input data $X_i$ is passed through the initialized autoencoder to obtain latent embeddings $E_i \in \mathbb{R}^{256}$. These embeddings serve as the basis for estimating initial cluster centroids $\{C_j\}_{j=1}^k$ using the $k$-means algorithm. The centroids are stored as trainable weights within the clustering layer, enabling subsequent refinement during training.

Formally, given an input latent vector $E_i$, the clustering layer computes the similarity between $E_i$ and each centroid $C_j$ using the Student’s $t$-distribution, which provides a soft assignment $q_{ij}$ of sample $i$ to cluster $j$:
\[
q_{tj} \;=\; \frac{ \left(1 + \|E_t - C_j\|^2 / \alpha \right)^{-\frac{\alpha+1}{2}} }
{ \sum_{\ell=1}^k \left(1 + \|E_t - C_\ell\|^2 / \alpha \right)^{-\frac{\alpha+1}{2}} },
\quad \text{with } \alpha=1.
\]
where $\alpha$ is the degree of freedom of the distribution, set to $\alpha = 1$ in this work. The quantity $q_{ij}$ thus represents the probability of assigning latent vector $E_i$ to cluster $j$. This heavy-tailed distribution, inspired by t-SNE, prevents the dominance of large distances and ensures that close samples receive higher confidence assignments.

To stabilize training and emphasize confident assignments, an auxiliary target distribution $p_{ij}$ is constructed from $q_{ij}$ at each update step:
\[
p_{ij} = \frac{q_{ij}^2 / \sum_i q_{ij}}
{\sum\limits_{l=1}^{k} \left( q_{il}^2 / \sum_i q_{il} \right)}.
\]
This target distribution amplifies high-confidence assignments while reducing the effect of uncertain ones, preventing centroid collapse and encouraging cluster separation. During optimization, both the cluster centroids $\{C_j\}$ and the latent embeddings $\{E_i\}$ are refined using gradient descent by minimizing the Kullback–Leibler (KL) divergence between the soft assignments $Q = \{q_{ij}\}$ and the target distribution $P = \{p_{ij}\}$:
\[
\mathcal{L}_{\text{clus}} = \frac{1}{T} \sum_{i=1}^{T} \sum_{j=1}^{k} p_{ij} \log \frac{p_{ij}}{q_{ij}}.
\]

This iterative process alternates between (i) updating $p_{ij}$ from the current $q_{ij}$ distribution and (ii) refining the centroids and latent mappings via backpropagation. The procedure continues until cluster assignments converge, measured by the stability of predicted labels across iterations. If the proportion of changing assignments falls below a threshold tolerance for multiple consecutive iterations, training is terminated, and the final cluster assignments $\{C_j\}_{j=1}^k$ are returned.

The implementation integrates this clustering layer into the autoencoder model, outputting both the reconstructed sequence and the cluster probabilities. By coupling the clustering objective with the reconstruction loss, the model learns discriminative latent embeddings that not only preserve the spatiotemporal dynamics of the original dataset but also yield meaningful and well-separated clusters.

\subsection{Joint Optimization of Reconstruction and Clustering}

The proposed training procedure jointly optimizes the autoencoder parameters and the clustering assignments by minimizing the mean squared error and the Kullback-Leibler (KL) divergence loss respectively so that the latent space becomes both information-preserving and cluster-separable. This optimization task is implemented using the Stochastic Gradient Descent (SGD) method with momentum. SGD guides the autoencoder model to learn efficient latent embeddings to capture distinctive and representative features of the input dataset. Let $\{X_t\}_{t=1}^{T}$ be the input sequence and $\{\widehat{X}_t\}_{t=1}^{T}$ its reconstruction. Let $E_t \in \mathbb{R}^m$ denote the latent embedding of $X_t$ produced by the encoder, and let $\{C_j\}_{j=1}^{k}$ be the learnable cluster centroids in the latent space (realized as a trainable weight matrix in the \textit{ClusteringLayer}). Training proceeds by minimizing a composite loss that balances a reconstruction term and a clustering term. The reconstruction term enforces fidelity between inputs and outputs and encourages the encoder-decoder to retain physically meaningful spatiotemporal structure:
\begin{equation}
\mathcal{L}_{\mathrm{rec}} \;=\; \frac{1}{T} \sum_{t=1}^{T} \big\| X_t - \widehat{X}_t \big\|_2^2,
\quad
X_t, \widehat{X}_t \in \mathbb{R}^{\mathrm{lon}\times \mathrm{lat}\times n}.
\label{eq:recon}
\end{equation}
This term is realized as mean squared error (MSE) in the implementation, attached to the autoencoder output head. To transform the latent space into well-separated regimes, we adopt a Student's $t$-kernel to compute soft assignments from the latent embedding $E_t$ to centroids $\{C_j\}_{j=1}^{k}$:
Following deep embedded clustering (DEC), we refine the assignments using a sharpened \emph{target distribution} that emphasizes confident predictions and balances centroid contributions:
The clustering objective minimizes the Kullback–Leibler divergence between the target and the current soft assignments. In the implementation, $\mathcal{L}_{\mathrm{clus}}$ is attached to the clustering head; the \textit{ClusteringLayer} exposes $q=\{q_{tj}\}$ and the training loop periodically recomputes $p=\{p_{tj}\}$ via \textit{target\_distribution}.

The overall objective balances faithful reconstruction with discriminative clustering:
\begin{equation}
\mathcal{L}_{\mathrm{total}} 
\;=\; \mathcal{L}_{\mathrm{rec}} \;+\; \lambda\, \mathcal{L}_{\mathrm{clus}},
\qquad \lambda>0,
\label{eq:total}
\end{equation}
implemented with loss weights \textit{['mse','kld']} on the two output heads (the scalar $\lambda$ can be realized via Keras loss weights if desired). The network is optimized end-to-end using stochastic gradient descent with momentum:
\begin{equation}
\theta \leftarrow \theta \;-\; \eta\, \nabla_{\theta} \mathcal{L}_{\mathrm{total}} 
\;+\; \mu\, (\theta - \theta_{\mathrm{prev}}),
\end{equation}
where $\eta$ is the learning rate and $\mu$ is the momentum coefficient. This jointly updates the encoder–decoder weights, the clustering head, and the centroid parameters $\{C_j\}$.
To stabilize optimization, cluster centroids are initialized by $k$-means on the encoder features $\{E_t\}$: \(\{C_j\}_{j=1}^{k} \;\leftarrow\; \mathrm{kmeans}\big(\{E_t\}_{t=1}^{T}\big).\)
Training then alternates between mini-batch gradient updates and periodic target refreshes. Every \textit{update\_interval} iterations, the current soft assignments $q$ are computed on the full dataset, the target distribution $p$ is updated, and the fraction of label changes $\Delta = \frac{1}{T}\sum_{t}\mathbb{1}\{\arg\max_j q_{tj}\neq \arg\max_j q^{\mathrm{prev}}_{tj}\}$ is monitored. When $\Delta < \textit{tol}$, training halts (an optional patience can be used to require multiple consecutive satisfactions).

At each training step with batch $\mathcal{B}$, the model minimizes
\[
\mathcal{L}_{\mathrm{batch}} \;=\; 
\frac{1}{|\mathcal{B}|}\sum_{t\in\mathcal{B}} \big\| V_t - \widehat{V}_t \big\|_2^2
\;+\; \lambda\, \frac{1}{|\mathcal{B}|}\sum_{t\in\mathcal{B}} \sum_{j=1}^{k} 
p_{tj}\,\log\!\frac{p_{tj}}{q_{tj}},
\]
where $p_{tj}$ are taken from the most recent target refresh. Periodic checkpointing persists model weights for reproducibility.

The MSE term in Equation \eqref{eq:recon} prevents the latent space from collapsing by requiring that $E_t$ preserve information sufficient for high-fidelity reconstruction, while the KL term sculpts the latent geometry to concentrate samples into well-separated, temporally coherent regimes. Their combination \eqref{eq:total} yields embeddings that are simultaneously compact, physically faithful, and cluster-discriminative. 
Training optimizes a composite loss that couples faithful reconstruction with discriminative clustering. The reconstruction term measures spatiotemporal fidelity via mean squared error,

The proposed architecture unifies multiscale ConvLSTM2D encoding, U-Net detail preservation, and B\text{-}TGAT temporal abstraction within a single end-to-end trainable system. By combining a principled clustering objective with faithful sequence reconstruction, the method learns latent embeddings that expose temporally coherent climate regimes and support robust, interpretable clustering across complex, nonlinear, multivariate spatiotemporal datasets.

\section{Experiment}\label{sec:exp}
All models are executed on AWS cloud environment using 20GB of S3 storage with 30 GB of ml.g4dn.xlarge GPU. The hardware used is a macOS Sonoma version 14.4.1, 16 GB, M1 pro chip. We applied the same python library across all models for homogeneity. 


\subsection{Dataset and Data Preprocessing}\label{ch:preprocesss}
This section introduces all datasets used in this research. To ensure generalizability, we experimented with three distinct multivariate spatiotemporal datasets.\\

\noindent \textit{C3S Arctic Regional Reanalysis (CARRA)}

\begin{table}[H]
\caption{CARRA - Data Description.}
\begin{center}
\label{tab:my-table3}
\setlength\tabcolsep{7pt}
\begin{tabular}{|l|l|l|l|}
\hline
\textbf{Var} & \textbf{Variable}         & \textbf{Range}                    & \textbf{Unit}             \\ \hline
tp           & Total precipitation       & {[}0, 0.0014{]}            & \(m\)    \\ \hline
rsn          & Snow density              & {[}99.9, 439.9{]}         & \(kg/m^3\)  \\ \hline
strd & Surface long-wave& {[}352599.1, 1232191.0{]} & \(J/m^2\) \\ \hline
t2m          & 2m  temperature           & {[}224.5, 289.4{]}        & \(k\)    \\ \hline
smlt         & Snowmelt                  & {[}-2.9\(e^{11}\),  8.5\(e^{04}\) {]} & \(m\)       \\ \hline
skt          & Skin temperature          & {[}216.6, 293.3{]}        & k \\ \hline
u10          & 10m u-wind & {[}-9.4, 13.3{]}        & \(m/s\)   \\ \hline
v10          & 10m v-wind& {[}-22.4, 16.1{]}       & \(m/s\)  \\ \hline
tcc          & Total cloud cover         & {[}0.0, 1.0{]} & \((0 - 1)\) \\ \hline
sd           & Snow depth                & {[}0, 6.5{]} & \(m\)      \\ \hline
msl          & Mean sea level pres   & {[}97282.1, 105330.8{]}  & \(Pa\)     \\ \hline
ssrd & Surface short-wave & {[}0, 1670912.0{]}      & \(J/m^2\) \\ \hline
\end{tabular}
\end{center}
\end{table} The C3S Arctic Regional Reanalysis (CARRA) dataset contains 3-hourly analyses and hourly short term forecasts of atmospheric and surface meteorological variables at 2.5 km resolution~\cite{CARRA}. Table \ref{tab:my-table3} presents a list of variables and ranges of daily observation over a period of one year. The data is 4-dimensional: longitude, latitude, time and variables (\(8 \times 18 \times 365 \times 13\)) respectively. After preprocessing, we get a 2D high dimensional data with dimensions (\(365 \times 29,952\)).
\subsection{ERA-5 Global Reanalysis}
\begin{table}[H]
\caption{ERA-5 Data Description.}
\begin{center}
\label{tab:my-table1}
\setlength\tabcolsep{3pt}
\begin{tabular}{|l|l|l|l|}
        \hline
        \textbf{Var} & \textbf{Variable}          & \textbf{Range}         & \textbf{Unit} \\ \hline
        sst          & Sea Surface Temperature    & [285, 300]           & \(k\)\\ \hline
        slp          & Sea Level Pressure         & [98260, 103788]    & \(pa\)  \\ \hline
        sshf         & Surface Sensible Heat Flux & [-674528, 200024]  & \(J/m^2\) \\ \hline
        t2m          & 2m  temperature            & [281, 
 299]           & \(k\)  \\ \hline
        slhf         & Surface Latent Heat Flux   & [-1840906, 90131] & \(J/m^2\)\\ \hline
        u10          & 10m u-wind component  & [-16, 19]            & \(m/s\) \\ \hline
        v10          & 10m v-wind component  & [-15, 16]              & \(m/s\)  \\ \hline
        \end{tabular}
    \end{center}
\end{table}
Table~\ref{tab:my-table1} lists the variables selected from the open-access ERA5 global atmospheric reanalysis produced by the European Centre for Medium-Range Weather Forecasts (ECMWF)~\cite{ERA5}. ERA5 provides a consistent, detailed record of the Earth’s atmosphere, surface, and ocean by assimilating diverse observations into a modern numerical weather prediction system; it offers about 31\,km horizontal resolution for the deterministic reanalysis and a reduced-resolution ten-member ensemble~\cite{hersbach2020era5}. Table~\ref{tab:my-table1} summarizes each variable and its physical range. These variables were chosen for their relevance to air--sea--cloud interactions. The dataset used here comprises daily fields over one year on a \(41\times 41\) latitude--longitude grid, with seven variables per grid cell \([41\times 41\times 7]\).

\subsection{NCEP/NCAR Reanalysis 1}
\begin{table}[H]
\caption{NCEP/NCAR Reanalysis 1 - Data Description.}
\begin{center}
\label{tab:my-table4}
\begin{tabular}{|l|l|l|}
        \hline
        \textbf{Var} & \textbf{Variable}& \textbf{Unit} \\ \hline
        climatology\_bounds & Climate Time Boundaries  & \(Hours\)            \\ \hline
        Air         & Air temperature at 2 m & \(Deg K \)          \\ 
        \hline
        valid\_yr\_count & non-missing values & \(Int\)          \\ \hline
        \end{tabular}
    \end{center}
\end{table} Daily atmospheric observations for calendar year 2019 were interpolated to standard pressure surfaces and mapped to a global $2.5^\circ \times 2.5^\circ$ latitude–longitude grid ($144 \times 73$ cells). Table~\ref{tab:my-table4} lists the variables retained, selected for their relevance to long-term climate variability and historical weather analysis.
For downstream modeling, we use two complementary representations: (i) a 4D tensor with explicit spatial and channel axes, $\mathbb{R}^{365\times 94\times 192\times 3}$ (time $\times$  lon $\times$ lat  $\times$ variables), and (ii) a 2D matrix obtained by flattening the spatial–channel dimensions across time, $\mathbb{R}^{365\times 108{,}288}$.
\subsection{Data Preprocessing}
All datasets comprise daily 4D spatiotemporal arrays over one year with dimensions \([\text{lon},\,\text{lat},\,\text{time},\,\text{variables}]\). To interface with clustering methods that require 2D inputs, we reshape each dataset to a high-dimensional matrix \([\text{time},\,\text{lon}\times\text{lat}\times\text{variables})]\). Missing values attributed to sensor gaps or adverse conditions are imputed using the overall mean to preserve temporal continuity. The null values are replaced by the overall mean of the dataset so as not to obstruct the temporal pattern, which obviously will change the actual behavior of variable in the dataset. Finally, all features are scaled via Min–Max normalization to the \([0,1]\) range to ensure numerical stability across variables. We used the distortion score elbow approach to determine the optimal number of clusters present in each dataset.

\begin{figure}[H]
  \centering
  \includegraphics[width=\linewidth]{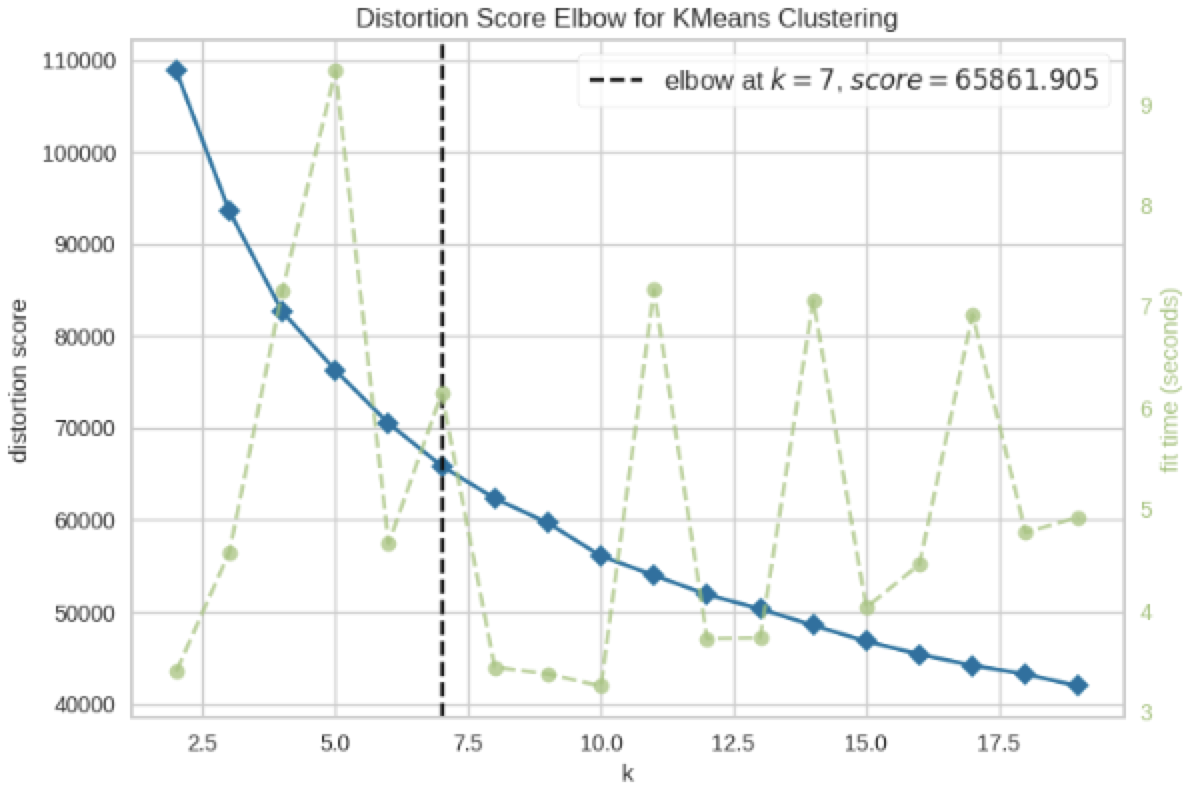}
  \centering \caption{Optimal Number of Clusters}
  \label{fig:opt_clu}
\end{figure}
From Figure \ref{fig:opt_clu} 
we can conclude that the optimal number of clusters is 7. To further confirm our optimal number of clusters, we tested K = 6, K = 7, and K = 8 and had better results at K = 7 

\begin{table*}[htp]
\begin{center}
\caption{Performance evaluation of our proposed model: We compare B\text{-}TGAT against a mixture of traditional and deep learning clustering models on three spatiotemporal dataset using six internal cluster evaluation metrics.}
\label{tab:my-finalP}
\resizebox{0.7\textwidth}{!}{%
\begin{tabular}{lc|ccccc|c|}
\cline{3-8}
 &
  \multicolumn{1}{l|}{} &
  \multicolumn{5}{c|}{\textbf{Baseline Models}} &
  \textbf{Ours} \\ \hline
\multicolumn{1}{|c|}{\multirow{7}{*}{\begin{tabular}[c]{@{}c@{}}ERA5\\ \\ \\ \\ 7 optimal\\ clusters\end{tabular}}} &
  \textbf{Performance} &
  \multicolumn{1}{c|}{\textbf{KMeans}} &
  \multicolumn{1}{c|}{\textbf{DTC}} &
  \multicolumn{1}{c|}{\textbf{DSC}} &
  \multicolumn{1}{c|}{\textbf{DEC}} &
  \textbf{HAC} &
  \textbf{B\text{-}TGAT} \\ \cline{2-8} 
\multicolumn{1}{|c|}{} &
  \textbf{Silhouette \(\uparrow\)} &
  \multicolumn{1}{c|}{0.2664} &
  \multicolumn{1}{c|}{0.3023} &
  \multicolumn{1}{c|}{0.3195} &
  \multicolumn{1}{c|}{0.3135} &
  0.2354 & \textbf{0.3268}
  \\ \cline{2-8} 
\multicolumn{1}{|c|}{} &
  \textbf{DB \(\downarrow\)} &
  \multicolumn{1}{c|}{1.6102} &
  \multicolumn{1}{c|}{1.5294} &
  \multicolumn{1}{c|}{1.6319} &
  \multicolumn{1}{c|}{1.5144} &
  1.5669 &\textbf{1.5009}
   \\ \cline{2-8} 
\multicolumn{1}{|c|}{} &
  \textbf{CH \(\uparrow\)}  &
  \multicolumn{1}{c|}{106.4115} &
  \multicolumn{1}{c|}{89.3054} &
  \multicolumn{1}{c|}{98.5739} &
  \multicolumn{1}{c|}{103.5368} &
  97.6476 &82.8211
  \\ \cline{2-8} 
\multicolumn{1}{|c|}{} &
  \textbf{RMSE \(\downarrow\)} &
  \multicolumn{1}{c|}{13.5104} &
  \multicolumn{1}{c|}{14.1963} &
  \multicolumn{1}{c|}{13.7743} &
  \multicolumn{1}{c|}{13.5633} &
  13.8148 &\textbf{13.2158}
 \\ \cline{2-8} 
\multicolumn{1}{|c|}{} &
  \textbf{Variance \(\downarrow\)} &
  \multicolumn{1}{c|}{0.1032} &
  \multicolumn{1}{c|}{0.04511} &
  \multicolumn{1}{c|}{0.1034} &
  \multicolumn{1}{c|}{0.0452} &
  0.1032 &0.1038
 \\ \cline{2-8} 
\multicolumn{1}{|c|}{} &
  \textbf{I-CD \(\uparrow\)} &
  \multicolumn{1}{c|}{6.3878} &
  \multicolumn{1}{c|}{7.2847} &
  \multicolumn{1}{c|}{7.0952} &
  \multicolumn{1}{c|}{6.7415} &
  7.0794 &\textbf{7.4839}
   \\ \hline
\multicolumn{1}{|l|}{} &
  \multicolumn{1}{l|}{} &
  \multicolumn{1}{c|}{} &
  \multicolumn{1}{c|}{} &
  \multicolumn{1}{c|}{} &
  \multicolumn{1}{c|}{} &
   &
   \\ \hline
\multicolumn{1}{|l|}{\multirow{6}{*}{\begin{tabular}[c]{@{}l@{}}CARRA\\ \\ 5 optimal\\ clusters\end{tabular}}} &
  \textbf{Silhouette \(\uparrow\)} &
  \multicolumn{1}{c|}{0.1812} &
  \multicolumn{1}{c|}{0.2666} &
  \multicolumn{1}{c|}{0.2434} &
  \multicolumn{1}{c|}{0.1657} &
  0.2131 &\textbf{0.2767}
   \\ \cline{2-8} 
\multicolumn{1}{|l|}{} &
  \textbf{DB \(\downarrow\)} &
  \multicolumn{1}{c|}{2.4387} &
  \multicolumn{1}{c|}{1.5528} &
  \multicolumn{1}{c|}{1.9968} &
  \multicolumn{1}{c|}{2.1326} &
  1.7226 & \textbf{1.5089}
  \\ \cline{2-8} 
\multicolumn{1}{|l|}{} &
  \textbf{CH \(\uparrow\)} &
  \multicolumn{1}{c|}{51.0463} &
  \multicolumn{1}{c|}{77.7148} &
  \multicolumn{1}{c|}{60.5852} &
  \multicolumn{1}{c|}{56.8826} &
  64.5219 &69.7729
   \\ \cline{2-8} 
\multicolumn{1}{|l|}{} &
  \textbf{RMSE \(\downarrow\)} &
  \multicolumn{1}{c|}{39.9203} &
  \multicolumn{1}{c|}{5.8977} &
  \multicolumn{1}{c|}{5.6710} &
  \multicolumn{1}{c|}{6.3021} &
  5.8429 &\textbf{5.5424}
 \\ \cline{2-8} 
\multicolumn{1}{|l|}{} &
  \textbf{Variance \(\downarrow\)} &
  \multicolumn{1}{c|}{0.1021} &
  \multicolumn{1}{c|}{0.1020} &
  \multicolumn{1}{c|}{0.0110} &
  \multicolumn{1}{c|}{0.1023} &
  0.1021 &\textbf{0.0105}
   \\ \cline{2-8} 
\multicolumn{1}{|l|}{} &
  \textbf{I-CD \(\uparrow\)} &
  \multicolumn{1}{c|}{34.1970} &
  \multicolumn{1}{c|}{{ 3.4306}} &
  \multicolumn{1}{c|}{3.1062} &
  \multicolumn{1}{c|}{2.9173} &
  3.3412 &3.0912
  \\ \hline
\multicolumn{1}{|l|}{} &
  \multicolumn{1}{l|}{} &
  \multicolumn{1}{l|}{} &
  \multicolumn{1}{l|}{} &
  \multicolumn{1}{l|}{} &
  \multicolumn{1}{l|}{} &
  \multicolumn{1}{l|}{} &
  \multicolumn{1}{l|}{} \\ \hline
\multicolumn{1}{|l|}{\multirow{6}{*}{\begin{tabular}[c]{@{}l@{}}NCAR\\ Reanalysis 1\\ \\ 7 optimal\\ clusters\end{tabular}}} &
  \textbf{Silhouette \(\uparrow\)} &
  \multicolumn{1}{c|}{0.6285} &
  \multicolumn{1}{c|}{0.6230} &
  \multicolumn{1}{c|}{0.61563} &
  \multicolumn{1}{c|}{0.5314} &
  0.6169 &\textbf{0.6541}
   \\ \cline{2-8} 
\multicolumn{1}{|l|}{} &
  \textbf{DB \(\downarrow\)} &
  \multicolumn{1}{c|}{0.8177} &
  \multicolumn{1}{c|}{0.7570} &
  \multicolumn{1}{c|}{0.7804} &
  \multicolumn{1}{c|}{0.6519} &
  0.7508 &\textbf{0.7612}
   \\ \cline{2-8} 
\multicolumn{1}{|l|}{} &
  \textbf{CH \(\uparrow\)} &
  \multicolumn{1}{c|}{842.7037} &
  \multicolumn{1}{c|}{864.1750} &
  \multicolumn{1}{c|}{862.3665} &
  \multicolumn{1}{c|}{817.3811} &
  759.6760 &\textbf{868.7555}
  \\ \cline{2-8} 
\multicolumn{1}{|l|}{} &
  \textbf{RMSE \(\downarrow\)} &
  \multicolumn{1}{c|}{3.4598} &
  \multicolumn{1}{c|}{3.1380} &
  \multicolumn{1}{c|}{3.1410} &
  \multicolumn{1}{c|}{3.1223} &
  3.6284 &\textbf{3.118}
  \\ \cline{2-8} 
\multicolumn{1}{|l|}{} &
  \textbf{Variance \(\downarrow\)} &
  \multicolumn{1}{c|}{0.1770} &
  \multicolumn{1}{c|}{0.1770} &
  \multicolumn{1}{c|}{0.1770} &
  \multicolumn{1}{c|}{0.1770} &
  0.1770 &0.1770
  \\ \cline{2-8} 
\multicolumn{1}{|l|}{} &
  \textbf{I-CD \(\uparrow\)} &
  \multicolumn{1}{c|}{0.9093} &
  \multicolumn{1}{c|}{0.8603} &
  \multicolumn{1}{c|}{0.8745} &
  \multicolumn{1}{c|}{0.8190} &
  0.8674 &\textbf{0.9098}
   \\ \hline
\end{tabular}%
}
\end{center}
\end{table*}

\subsection{Baseline Methods} We compare our proposed model against state-of-the-art conventional and deep clustering models. These include KMeans \citep{sinaga2020unsupervisedKMeans}, 
Hierarchical agglomerative clustering \citep{murtagh2017algorithms}, deep embedded clustering (DEC) \citep{xie2016unsupervised}, Deep Spatio-temporal Clustering (DSC) \citep{faruque2023deep} and Deep Temporal Convolution (DTC) \citep{sai2018deep} respectively. Based on the elbow method, we used \(k = 7\) across all algorithms in our experiments.

\subsection{Evaluation Metrics}\label{sec:Evaluation}

In the absence of ground truth, we evaluate the performance of our proposed model on six internal cluster validation measures: Silhouette Score \citep{ rousseeuw1987silhouettes}, Davies-Bouldin score (DB) \citep{ros2023pdbi}, Calinski-Harabas score (CH) \citep{wang2019improved}, Average inter-cluster distance (I-CD) \citep{everitt2011cluster}, Average Variance (Variance) \citep{montgomery2010applied} and Average root mean squared error (RMSE) \citep{willmott2005advantages}. 
These measures seek to balance the \textit{compactness} and the \textit{separation} of formed clusters through minimizing intra-cluster distance and maximizing the inter-cluster distance respectively.

\subsection{Experiment Results}\label{tab:Results}  Table~\ref{tab:my-finalP} reports internal cluster–validation scores. B\text{-}TGAT consistently outperforms all baselines; on ERA5 in particular it achieves the best overall performance on four of six metrics. By conditioning each representation on both past and future context over a spatiotemporal graph with relative-time encodings, B\text{-}TGAT captures full event morphology (onset$\rightarrow$peak$\rightarrow$decay), variable lead–lag relationships, and long-range teleconnections that deep clustering baselines (DSC, DEC) lose by aggregating time naively. Furthermore, B\text{-}TGAT learns time-varying edge weights via multi-head attention, adapting to non-stationarity and regime shifts while denoising through context-aware message passing rather than fixed, hand-crafted affinities. Trained end-to-end with reconstruction/contrastive objectives and differentiable soft assignments, its bottleneck yields tighter, better-separated latent representations and stronger clustering under missing data, irregular sampling, and multiscale dynamics.

\section{Conclusion} \label{concl}

We propose a deep unsupervised Bi-directional Temporal Graph Attention Transformer (B\text{-}TGAT) framework for clustering 4D multidimensional multivariate spatiotemporal data without prior supervision. B\text{-}TGAT fuses local spatiotemporal feature extraction with global, time–aware message passing at the autoencoder bottleneck and then performs \emph{soft} clustering in the latent space. In practice, a u-net style ConvLSTM2D encoder preserves multiscale spatial and temporal structure, a B\text{-}TGAT layer aggregates information across temporally relevant neighbors via bi-directional multihead attention with time encodings, and a ConvLSTM2D decoder promotes information preservation through reconstruction. Cluster assignments are computed with a Student’s-\(t\) kernel, while a KL divergence objective sharpens cluster memberships in a fully differentiable manner. The joint loss couples reconstruction fidelity, cluster separability, and graph smoothness into a single end–to–end pipeline. 
Future work will pursue joint graph learning, richer uncertainty quantification for assignments, cross–domain transfer, and physics–guided constraints to further enhance robustness and interpretability.

\bibliographystyle{IEEEtran}
\bibliography{biblob}



\end{document}